\documentclass{article}

\usepackage{microtype}
\usepackage{graphicx}
\usepackage{subcaption}
\usepackage{booktabs} 

\usepackage{hyperref}

\usepackage[preprint]{icml2026}

\usepackage{amsmath}
\usepackage{amssymb}
\usepackage{mathtools}
\usepackage{amsthm}
\usepackage{algorithm} 
\usepackage{algorithmicx}
\usepackage{algpseudocode}
\usepackage{xurl}
\usepackage{pifont}
\usepackage{booktabs,tabularx}

\usepackage{xcolor} 
\usepackage[table]{xcolor} 
\usepackage{multirow}

\usepackage[capitalize,noabbrev]{cleveref}

\theoremstyle{plain}

\theoremstyle{definition}

\theoremstyle{remark}

\usepackage[textsize=tiny]{todonotes}

\icmltitlerunning{Rewards as Labels: Revisiting RLVR from a Classification Perspective}

\begin{document}

\twocolumn[
  \icmltitle{Rewards as Labels: Revisiting RLVR from a Classification Perspective}



  \icmlsetsymbol{equal}{*}
  \icmlsetsymbol{corresponding}{$\dagger$} 

  \begin{icmlauthorlist}
    \icmlauthor{Zepeng Zhai}{xhs,equal}
    \icmlauthor{Meilin Chen}{xhs,equal,corresponding}
    \icmlauthor{Jiaxuan Zhao}{xhs,iie,equal}
    \icmlauthor{Junlang Qian}{ntu}
    \icmlauthor{Lei Shen}{ict}
    \icmlauthor{Yuan Lu}{xhs,corresponding}

  \end{icmlauthorlist}

  \icmlaffiliation{xhs}{Xiaohongshu Inc.}
  \icmlaffiliation{iie}{Institute of Information Engineering, Chinese Academy of Sciences}
  \icmlaffiliation{ict}{Institute of Computing Technology, Chinese Academy of Sciences}
  \icmlaffiliation{ntu}{Nanyang Technological University}
  
  \icmlcorrespondingauthor{Meilin Chen}{chenmeilin2@xiaohongshu.com}
  \icmlcorrespondingauthor{Yuan Lu}{luyuan2@xiaohongshu.com}


  \vskip 0.3in
]



\printAffiliationsAndNotice{}  

\begin{abstract}
Reinforcement Learning with Verifiable Rewards has recently advanced the capabilities of Large Language Models for complex reasoning tasks by providing explicit rule-based supervision. Among RLVR methods, GRPO and its variants have achieved strong empirical performance. Despite their success, we identify that they suffer from Gradient Misassignment in Positives and Gradient Domination in Negatives, which leads to inefficient and suboptimal policy updates. To address these issues, we propose \textit{\textbf{\underline{Re}wards} \textbf{\underline{a}s} \textbf{\underline{L}abels}} (\textbf{REAL}), a novel framework that revisits verifiable rewards as categorical labels rather than scalar weights, thereby reformulating policy optimization as a classification problem. Building on this, we further introduce anchor logits to enhance policy learning. Our analysis reveals that REAL induces a monotonic and bounded gradient weighting, enabling balanced gradient allocation across rollouts and effectively mitigating the identified mismatches. 
Extensive experiments on mathematical reasoning benchmarks show that REAL improves training stability and consistently outperforms GRPO and strong variants such as DAPO. On the 1.5B model, REAL improves average Pass@1 over DAPO by 6.7\%. These gains further scale to 7B model, where REAL continues to outperform DAPO and GSPO by 6.2\% and 1.7\%, respectively. Notably, even with a vanilla binary cross-entropy, REAL remains stable and exceeds DAPO by 4.5\% on average.
\end{abstract}

\vspace{-2em}

\begin{figure}[t]
\centering
\includegraphics[width=1\columnwidth]{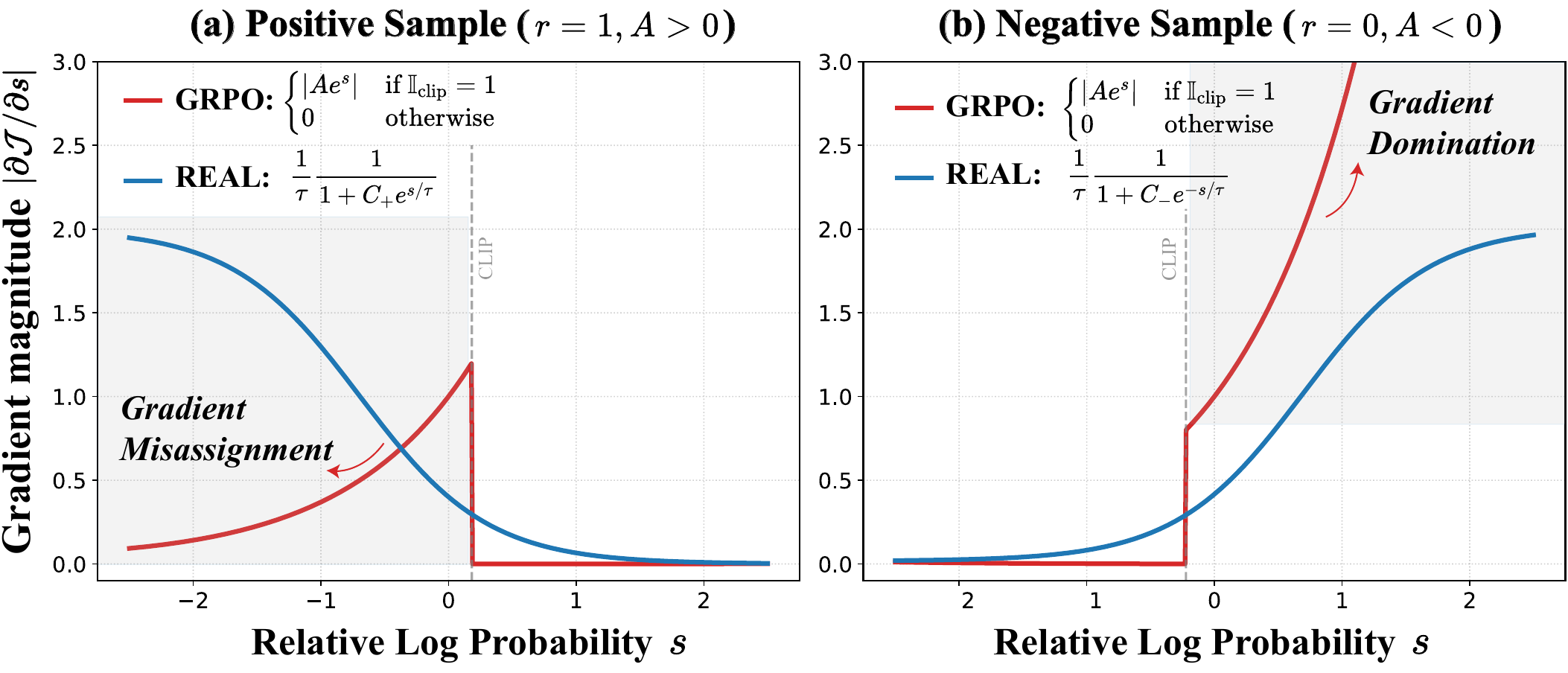}
\caption{\textbf{Gradient magnitude visualizations in GRPO and proposed REAL.} We visualize the gradient magnitude $|\mathcal{W}_\text{GRPO}|$ against relative log-probability $s_t$  (\textcolor{red}{shown in red}) and $|\mathcal{W}_\text{REAL}|$ against length-normalized relative log probability score $\bar s$ (\textcolor{blue}{shown in blue}) . \textbf{(a) Positive samples ($r=1$):} GRPO suffers from \textit{Gradient Misassignment}. In contrast, the gradient magnitude of our proposed REAL monotonically decreases with increasing relative log-probability. \textbf{(b) Negative samples ($r=0$):} GRPO exhibits \textit{Gradient Domination}, whereas REAL enforces strictly bounded gradients, ensuring stable training. For the curves shown in the figure, GRPO uses clipping $\varepsilon = 0.2$, and $A = 1$ as defined in Eq~ \ref{eq:grpo gradient},  while REAL uses temperature $\tau = 0.5$ and $C_+ = C_- = 4$ as defined in Eq~ \ref{eq:real gradient}.}

\vspace{-1.6em}
\label{fig:intro-1}
\end{figure}

\section{Introduction}
Reinforcement Learning with Verifiable Rewards (RLVR) has emerged as an effective post-training paradigm for improving large language models on tasks with rule-based evaluation, such as mathematical and program reasoning. Unlike Reinforcement Learning from Human Feedback (RLHF), RLVR uses verifiable reward functions to provide explicit supervision without relying on human preference judgments or learned reward models.

Group Relative Policy Optimization (GRPO)~\cite{shao2024deepseekmath} is a representative RLVR method. It normalizes rewards across multiple rollouts from the same prompt, which stabilizes updates and has been adopted in strong reasoning systems such as DeepSeek-R1~\cite{guo2025deepseek}. Building on GRPO, several variants aim to further improve training stability and efficiency, including DAPO~\cite{yu2025dapo}, GSPO~\cite{zheng2025group}, and related methods~\cite{chu2025gpg, lin2025cppo, su2025trust}.

Despite these empirical successes, Fig.~\ref{fig:intro-1} suggests that the policy gradient induced by GRPO-style methods exhibits two fundamental mismatches. These issues are especially pronounced for \emph{hard tokens}: \emph{Gradient Misassignment in Positives}, which under-updates low-confidence positive tokens, and \emph{Gradient Domination in Negatives}, which over-weights high-confidence negative tokens and allows them to dominate the update.
\vspace{-0.6em}
\begin{itemize}
\item  \textbf{Gradient Misassignment in Positives}. For positive rollouts, tokens that already have high probabilities under the current policy receive disproportionately large updates, while low-probability (\emph{hard}) tokens are assigned much weaker gradients. This biased weighting concentrates gradient mass on already well-optimized regions of the policy rather than correcting under-optimized components.
\item \textbf{Gradient Domination in Negatives}. For negative rollouts, gradient magnitudes are unbounded, causing high-probability (\emph{hard}) tokens to dominate the update and overwhelm contributions from lower-probability tokens, leading to imbalanced credit assignment.

\end{itemize}
\vspace{-0.6em}
These gradient mismatches reveal a fundamental misalignment between the desired credit assignment dictated by verifiable rewards and the actual gradient allocation induced by GRPO-style objectives. As a result, credit is assigned inefficiently across the policy space, which not only hampers effective policy improvement but also increases the risk of premature convergence to suboptimal local optima.

To address these issues, we reconsider the principle of policy optimization from a different perspective. Rather than treating verifiable rewards as scalar signals for policy gradient weighting, we reconceptualize them as coarse category labels. Under this view, policy optimization can be naturally reformulated as a classification task, where the objective is to correctly discriminate between desirable and undesirable rollouts based on verifiable criteria.

Motivated by this insight, we propose a novel RLVR framework,  \textit{\textbf{\underline{Re}wards} \textbf{\underline{a}s} \textbf{\underline{L}abels}} (\textbf{REAL}), which formulates policy learning as a classification objective. Instead of treating verifiable rewards as scalar weights, REAL uses them as categorical supervision, encouraging the policy to increase the relative log-probabilities of positive rollouts and decrease those of negative ones (Fig.~\ref{fig:intro-1}). Building on this, we further introduce anchor logits to enhance policy learning. A theoretical analysis shows that REAL induces \emph{\textbf{monotonic}} and \emph{\textbf{bounded}} gradient magnitude, which regulates how gradient mass is allocated across rollouts, ensuring that low-probability positives are not under-updated and that a small number of negatives do not dominate the parameter update. Consequently, REAL mitigates both Gradient Misassignment in Positives and Gradient Domination in Negatives, enabling more balanced credit assignment and improved training stability.

Extensive experiments across diverse mathematical reasoning benchmarks (AIME 2024/2025, MATH 500, AMC 2023, Minerva, and Olympiad Bench) and model scales demonstrate that REAL improves training stability, and enhances performance over GRPO and its strong variants. Specifically, on the 1.5B model, REAL improves average Pass@1 over DAPO by 6.7\%. These gains further scale to 7B model, where REAL continues to outperform DAPO and GSPO by 6.2\% and 1.7\%, respectively. Moreover, due to the bounded gradient magnitude, REAL stays stable even without an explicit KL penalty and achieves competitive performance against strong recent baselines. Remarkably, even with a vanilla binary cross-entropy loss, REAL remains stable and outperforms DAPO by 4.5\% on average. 

In summary, our main contributions are as follows:
\vspace{-1em}
\begin{itemize} 
\item \textbf{Identification of fundamental issues in GRPO-style RLVR.}
We identify two gradient mismatches, \textit{Gradient Misassignment in Positives} and \textit{Gradient Domination in Negatives}, which hinder efficient credit assignment and lead to suboptimal policy updates.  
\item \textbf{Proposal of the \textit{\textbf{Re}wards \textbf{a}s \textbf{L}abels} (REAL) framework.}
By reframing verifiable rewards as categorical labels instead of scalar weights, REAL formulates policy optimization as a classification task, effectively mitigating the identified issues in GRPO-style RLVR.  
\item \textbf{Comprehensive empirical validation.} Extensive experiments on diverse mathematical reasoning benchmarks demonstrate that REAL consistently outperforms GRPO and its variants in training stability and reasoning performance.
\vspace{-1em}
\end{itemize}

\section{Preliminaries}

\subsection{Group Relative Policy Optimization}

Consider a model $\pi_\theta$ parameterized by $\theta$, and denote the model from the previous step by $\pi_{\text{old}}$. Given a prompt $q$, the generated output $o$ follows the distribution $\pi_{\text{old}}(\cdot\mid q)$, which includes reasoning traces and the final answer. Specifically, the output $o$ is generated token by token, i.e., 
\begin{align*}
o_t &\sim \pi_{\text{old}}\big(\cdot \mid q, o_{<t}\big), \quad t = 1, 2, \dots, |o|
\end{align*}
Group Relative Policy Optimization (GRPO) samples $G$ rollouts with rewards ${\{r^i\}_{i=1}^G}$ for advantage estimation:
\begin{align*}
\hat{A}_t^i = \frac{r^i - \text{mean}(\{r^i\}_{i=1}^G)}{\text{std}(\{r^i\}_{i=1}^G)}
\end{align*}
The loss function of GRPO is defined as:
\begin{align}
    & \mathcal{J}_{\text{GRPO}}(\theta) = \mathbb{E}_{q, o \sim \pi_{{\text{old}}}(\cdot|q)}  \notag \\
    & \Bigg[ \frac{1}{|o|} \sum_{t=1}^{|o|} \bigg( \min \Big( \rho^\theta_t {A}_t, \text{clip} \left( \rho^\theta_t, 1 - \varepsilon, 1 + \varepsilon \right) {A}_t \Big) \notag \\
    &\quad - \beta \mathbb{D}_{\text{KL}}(\pi_\theta \| \pi_{\text{ref}}) \bigg) \Bigg],
\end{align}
where $\rho^\theta_t = \frac{\pi_\theta(o_t|q)}{\pi_{\text{old}}(o_t|q)}$ is the Importance Sampling (IS) ratio, $\beta$ is a weight for the Kullback-Leibler (KL) divergence between the current policy $\pi_\theta$ and the reference policy $\pi_{\text{ref}}$, $A_t$ is the advantage and $\varepsilon$ is the clipping hyperparameter.


\begin{table}[t]
    \centering
    \caption{Percentage and Avg. Gradient Magnitudes with respect to the log-probability of tokens in different clipping range. The clip ratio is $\epsilon=0.2$. Zero gradients (indicated by -) occur where clipping is active.}
    \label{tab:grpo_stats}
    \resizebox{0.48\textwidth}{!}{%
    \begin{tabular}{llcc}
        \toprule
         & \textbf{Ratio Range} & \textbf{Pct. (\%)} & \textbf{Avg. Grad. Mag.} \\
        \midrule
        
        \multirow{4}{*}{\textbf{Positives}} 
          & $< 1-\varepsilon$                          & 0.22  & 0.6943 \\
          & $[1-\varepsilon,\; 1)$                     & 4.83  & 0.9788 \\
          & $[1,\; 1+\varepsilon]$                     & 94.63 & 1.0010 \\
          & $> 1+\varepsilon$ \scriptsize{(Clipped)}   & 0.32  & - \\
        \midrule
        
        \multirow{4}{*}{\textbf{Negatives}} 
          & $< 1-\varepsilon$ \scriptsize{(Clipped)}   & 0.24  & - \\
          & $[1-\varepsilon,\; 1)$                     & 5.47  & 0.9774 \\
          & $[1,\; 1+\varepsilon]$                     & 93.95 & 1.0012 \\
          & $> 1+\varepsilon$                          & 0.34  & 1.3553 \\
        
        \bottomrule
    \end{tabular}%
    }
\end{table}

\subsection{Unified Version of Softmax Loss}
From a classification perspective, we seek to separate positive from negative samples by assigning higher logits to positives. We therefore adopt a softmax cross-entropy objective to enforce this separation.
Let $\mathcal{Z}_p = \{z_p^{i}\}_{i=1}^{P}$ and $\mathcal{Z}_n = \{z_n^{j}\}_{j=1}^{N}$ denote the logits produced for positive and negative samples, respectively. Following \citet{sun2020circle} and \citet{su2022zlpr}, the standard softmax cross-entropy loss can be reformulated in the following unified form:
\vspace{-0.5em}
\begin{equation}
\label{eq:softmax loss}
\begin{aligned}
\mathcal{L}_{CE}(\mathcal{Z}_p, \mathcal{Z}_n)
&= \log \left( 1 + \sum_{i=1}^{P} \sum_{j=1}^{N} e^{(z_n^j - z_p^i)} \right) \\
&= \log \left( 1 + \sum_{j=1}^{N} e^{z_n^j} \sum_{i=1}^{P} e^{-z_p^i} \right).
\end{aligned}
\end{equation}
This formulation explicitly contrasts positive logits $\mathcal{Z}_p$ with negative logits $\mathcal{Z}_n$. For each pair $(z_p^i, z_n^j)$, the term $e^{z_n^j - z_p^i}$ encourages the model to increase the gap between the positive logit $z_p^i$ and the negative logit $z_n^j$. Minimizing this loss therefore raises positive scores while suppressing negative scores, increasing the margin between the two sets.

\section{Analysis of GRPO-Style Methods}

\textbf{Proposition 3.1.}
(\emph{refer to Appendix~\ref{appendix:Gradient Analysis of GRPO} for detailed proof}) 
GRPO induces an \emph{\textbf{unbounded}} gradient weighting over rollouts, where the contribution of individual tokens grows \emph{exponentially} with their relative log-probabilities. Concretely, for the $t$-th token of a rollout $o$, the magnitude of the GRPO gradient is given by:
\begin{equation}
\label{eq:grpo gradient}
|\mathcal{W}_\text{GRPO}| = 
\begin{cases} 
|A \cdot e^{s_t}|, & \text{if } \mathbb{I}_{\text{clip}}=1  \\
0, & \text{otherwise}.
\end{cases}
\end{equation}

where $\mathbb{I}_{clip}$ is the indicator function for the clipping constraint. $s_t$ is the relative log-probability between $\pi_\theta(o_t|q)$ and $\pi_{\text{old}}(o_t|q)$:
\begin{equation}
\label{eq:s_definition}
\begin{aligned}
s_t = \log \frac{\pi_\theta(o_t|q)}{\pi_{\text{old}}(o_t|q)}
\end{aligned}
\vspace{-1em}
\end{equation}

As illustrated in Fig.~\ref{fig:intro-1}, the magnitude of the GRPO gradient $|\mathcal{W}_\text{GRPO}|$ grows exponentially with the relative log-probability $s_t$, and this property inherently leads to two undesirable optimization behaviors.

\textbf{Gradient Misassignment in Positives}.
For positive rollouts, the gradient magnitude is proportional to $e^{s_t}$ and therefore decreases as the policy probability becomes smaller relative to the reference policy. This behavior is counterintuitive: tokens that the current policy underestimates should receive stronger corrective gradients to facilitate rapid improvement. Instead, GRPO assigns progressively weaker gradients to these hard positive tokens, resulting in insufficient optimization. Consequently, the learning signal for challenging yet correct outputs is significantly diminished.

\textbf{Gradient Domination in Negatives.}
In contrast, for negative rollouts, the gradient magnitude is not upper-bounded and grows exponentially with the relative log-probability $s_t$. Within the group-relative framework of GRPO, this leads to a severe imbalance: several negative samples with an excessively large relative log-probability can dominate the entire group gradient. As a result, the contributions of other informative negative samples are suppressed, making the optimization overly sensitive to outliers and reducing training stability.


These issues are most pronounced on \emph{hard tokens}, i.e., low-confidence positives and high-confidence negatives. Following the training settings in Sec.~\ref{sec:exp}, we analyze the statistical distribution of gradients with respect to the log-probability in Tab.~\ref{tab:grpo_stats}, which corroborates our theoretical analysis: for \textbf{hard positives} (ratio $< 1{-}\varepsilon$), the average gradient magnitude is suppressed to 0.6943, confirming that GRPO diminishes the learning signal where the policy is most under-trained. Meanwhile, 4.83\% of positive tokens fall in $[1{-}\varepsilon, 1)$ with a reduced magnitude of 0.9788, receiving weakened reinforcement despite belonging to correct responses. Conversely, for \textbf{hard negatives} (ratio $> 1{+}\varepsilon$), the magnitude surges to 1.3553, validating the tendency for outlier domination. This underscores that GRPO fails to assign appropriate gradient scales to these critical tokens, motivating the development of our proposed approach.

Importantly, unlike GRPO-style methods, REAL (as illustrated in the next section) does not require clipping: its bounded and monotonic gradient weighting naturally handles all rollouts and tokens, ensuring stable and effective optimization across the entire training distribution.

\begin{figure*}[t]
\centering 
\includegraphics[width=0.75\textwidth]{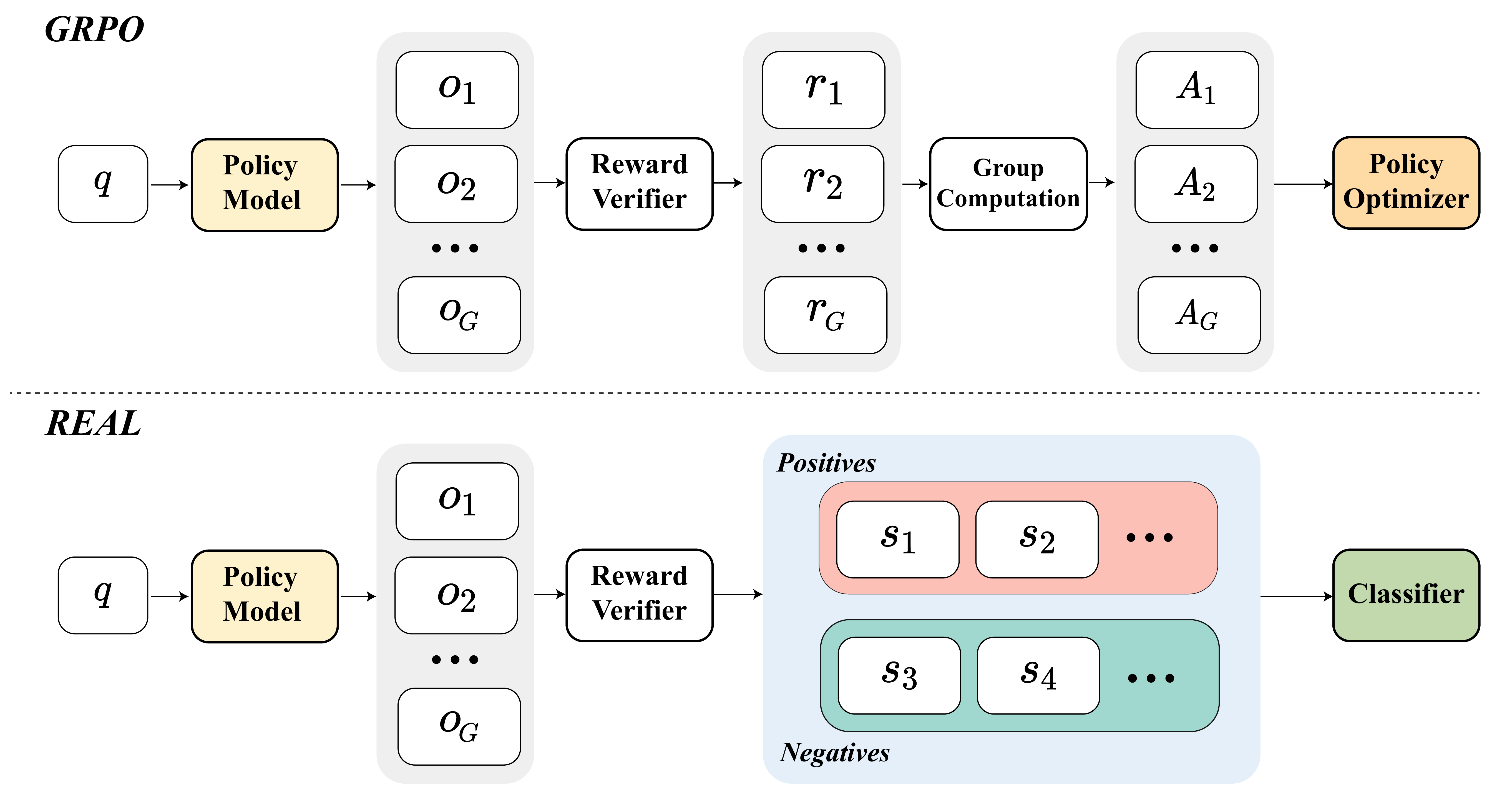}
\caption{Overview of our REAL framework. REAL formulates RLVR as a classification problem by treating rewards as explicit labels. $s$ denotes the prediction score.}
\vspace{-1em}
\label{figure.method.1}
\end{figure*}

\section{Rewards as Labels}
\subsection{Revisiting RLVR as Classification}

Consider the reinforcement learning with verifiable rewards (RLVR) setting. Given a prompt $q$, the current policy $\pi_\theta$ generates a group of $G$ rollouts
\[
\mathcal{O} = \{ o^1, \dots, o^G \},
\]
where each rollout $o^i$ is associated with a verifiable outcome reward
\[
r^i \in \{0, 1\},
\]
indicating whether the rollout satisfies a predefined verification rule.

\paragraph{Verifiable Rewards as Labels.}
We therefore reinterpret verifiable rewards not as scalar signals, but as \emph{coarse category labels} that partition rollouts into two disjoint sets:
\[
\mathcal{O}_+ = \{ o^i \mid r^i = 1 \}, \quad
\mathcal{O}_- = \{ o^j \mid r^j = 0 \}.
\]
Under this view, policy optimization could be reformulated as a classification problem that separates desired and undesirable rollouts within each prompt-conditioned group.

\paragraph{Relative Log-Prob as Logits.}
To operationalize this perspective, we define a logit score based on the policy update itself:
\[
\mathcal{S}_+ = \{ \bar{s}^i \mid r^i = 1 \}, \quad
\mathcal{S}_- = \{ \bar{s}^j \mid r^j = 0 \}.
\]
For $k$-th rollout $o^k$ in $\mathcal{S}_+$ or $\mathcal{S}_-$, we compute a length-normalized relative log probability score over its tokens as logits:
\begin{align}
\label{eq:relative log-prob}
\bar{s}^k
&=
\frac{1}{|o^k|}
\sum_{t=1}^{|o^k|}
\Bigl(
\log \frac{\pi_\theta(o_t^k \mid q)}
{\pi_{{\mathtt{old}}}(o_t^k \mid q)}
\Bigr) \notag \\
&=
\frac{1}{|o^k|}
\sum_{t=1}^{|o^k|}
\Bigl(
s_t^k
\Bigr)
\end{align}

which measures the relative change in the probability of the rollout under the new policy compared to the old policy.
This score has a clear interpretation: 
$ \bar{s}^k > 0$ indicates that the rollout is reinforced, while $ \bar{s}^k < 0$ indicates suppression.

\begin{algorithm}[t]
\caption{Rewards as Labels}
\begin{algorithmic}[1]
\State \textbf{Input:} initial policy model $\pi_0$, reward function $r(\cdot)$, task prompts $\mathcal{D}$, temperature $\tau$
\State policy model $\pi_\theta \leftarrow \pi_0$
\For{training step $k = 1$ to $T$}
    \State Sample a batch $\mathcal{B}$ from dataset $\mathcal{D}$
    \State Update the old policy model $\pi_{\text{old}} \leftarrow \pi_\theta$
    \State Sample $G$ outputs $\mathcal{O} = \{o^i\}_{i=1}^G \sim \pi_{\text{old}}(\cdot\mid q)$ for each question $q \in \mathcal{B}$
    \State Partition $\mathcal{O}$ into positive set $\mathcal{O}_+ = \{o^i : r(o^i|q) = 1\}$ and negative set $\mathcal{O}_- = \{o^j : r(o^j|q) = 0\}$

    \For{mini-batch $\mathcal{B}_k$ in $\mathcal{B}$}
        \State Calculate relative log-prob logits with Eq.~\ref{eq:relative log-prob}
        \State Update the policy model $\pi_{\theta}$ with the REAL objective (Eq.~\ref{eq:REAL})
    \EndFor
\EndFor
\State \textbf{Output:} Fine-tuned policy $\pi_\theta$
\end{algorithmic}
\label{algo:REAL}
\end{algorithm}

\subsection{REAL Objective}
\label{sec 5.2}

Building upon the above definitions, REAL directly optimizes a unified softmax objective:
\begin{align}
\label{eq:vanilla loss}
\mathcal{L}_{Vanilla} &= \mathcal{L}_{CE}(\mathcal{S}_+, \mathcal{S}_-) \notag 
\end{align}

Importantly, REAL constitutes a general optimization framework that can be naturally combined with a wide range of standard classification losses, such as binary cross-entropy (BCE). We provide a more detailed analysis in Sec.~\ref{sec:further analysis}.

\paragraph{Enhancement with Anchor Logits.} 
We introduce a fixed anchor logit at $0$, i.e., $\mathcal{S}_0 = \{\bar{s} = 0\}$, because our score function naturally uses $0$ as the threshold. Without this anchor, the objective mainly depends on the score gap between positives and negatives, which can lead to ambiguous update directions (e.g., the gap may increase even when both positive and negative scores decrease). For positive samples, the anchor $0$ is treated as a negative logit; for negative samples, it is treated as a positive logit. With the anchor, we enforce clear separation by increasing positive rollout scores above $0$ and decreasing negative rollout scores below $0$:

\begin{align}
\mathcal{L}_{REAL} &= \mathcal{L}_{CE}(\mathcal{S}_+, \mathcal{S}_0) + \mathcal{L}_{CE}(\mathcal{S}_0, \mathcal{S}_-) \notag 
\end{align}

Based on the softmax loss in Eq. \ref{eq:softmax loss}, the loss for positive rollouts can be formulated as:
\begin{align}
\mathcal{L}_+ &= \mathcal{L}_{CE}(\mathcal{S}_+, \mathcal{S}_0) \notag \\
& = \log (1 + \sum_{j=1}^{1} e^{0}
\sum_{\mathcal{O}_+} e^{-\bar{s}^i/\tau}) \notag \\
&= \log \left(
1 + \sum_{\mathcal{O}_+} e^{-\bar{s}^i/\tau}
\right),
\end{align}
where $\tau$ is a temperature parameter controlling the sharpness of the decision boundary.

Similarly, the loss for negative rollouts is defined as:
\begin{align}
\mathcal{L}_- &= \mathcal{L}_{CE}(\mathcal{S}_0, \mathcal{S}_-) \notag \\
&= \log \left(
1 + \sum_{\mathcal{O}_-} e^{\bar{s}^j/\tau}
\right)
\end{align}

By combining both terms, we obtain the final REAL loss:
\begin{align}
\label{eq:REAL}
& \mathcal{L}_{REAL} = \mathcal{L}_+ + \mathcal{L}_- \notag \\
& = \log \left(
1 + \sum_{\mathcal{O}_+} e^{-\bar{s}^i/\tau}
\right)
+
\log \left(
1 + \sum_{\mathcal{O}_-} e^{\bar{s}^j/\tau}
\right)
\end{align}

\paragraph{Discussion.}
The formulation in Eq.~\ref{eq:REAL} naturally balances gradient allocation across both positive and negative rollouts. For positive rollouts, a smaller $\bar{s}^i$ leads to a larger loss, thereby encouraging the policy to increase the log-probability of these desirable sequences. Conversely, for negative rollouts, a larger $\bar{s}^j$ increases the loss, pushing the policy to decrease the log-probability of undesirable sequences. This adaptive gradient modulation effectively mitigates the \textit{Gradient Misassignment in Positives} and \textit{Gradient Domination in Negatives} observed in previous approaches. Refer to Algorithm \ref{algo:REAL} for the detailed procedure for REAL.

\begin{table*}[t]
    \centering
    \caption{
    Performance comparison on the DeepScaleR-Preview-Dataset using the DeepSeek-R1-Distill-Qwen-1.5B backbone. OpenAI-o1-preview is included as a competitive reference. Results marked with * are cited from original reports or \citet{li2025disco}. REAL consistently outperforms all RLVR baselines across diverse benchmarks. DS, DSR is short for DeepSeek-R1 and DeepScaleR respectively.}

    \renewcommand{\arraystretch}{1.1}
    \setlength{\tabcolsep}{5.5pt}
    \small 
    
    \begin{tabular}{@{}l c c c c c c c@{}}
        \toprule
        \textbf{} & AIME24 & AIME25 & MATH500 & AMC23 & Minerva & O-Bench & Avg. \\
        \midrule
        
        \multicolumn{8}{l}{\textit{\textbf{Models}}} \\ 
        \hspace{1em} OpenAI-o1-Preview\textsuperscript{*} & 0.400 & -- & 0.814 & -- & -- & -- & -- \\
        \hspace{1em} DS-Distill-Qwen-1.5B\textsuperscript{*} & 0.181 & 0.215 & 0.758 & 0.515 & 0.237 & 0.353 & 0.376 \\
        \hspace{1em} DSR-1.5B-Preview\textsuperscript{*} & 0.358 & 0.258 & 0.860 & 0.679 & 0.297 & 0.473 & 0.488 \\

        \addlinespace[0.5ex] 
        \midrule
        
        \multicolumn{8}{l}{\textit{\textbf{Methods}}} \\
        \hspace{1em} GRPO & 0.235 & 0.229 & 0.824 & 0.600 & 0.264 & 0.435 & 0.431 \\
        \hspace{1em} DAPO & 0.298 & 0.215 & 0.837 & 0.663 & 0.295 & 0.448 & 0.459 \\
        \hspace{1em} TRPA\textsuperscript{*} & 0.354 & 0.235 & 0.835 & 0.653 & 0.283 & 0.458 & 0.470 \\
        \hspace{1em} GSPO & 0.377 & 0.296 & 0.877 & 0.730 & \textbf{0.326} & 0.506 & 0.519 \\
        \hspace{1em} REAL & \textbf{0.406} & \textbf{0.321} & \textbf{0.877} & \textbf{0.734} & 0.310 & \textbf{0.508} & \textbf{0.526} \\
        \bottomrule
    \end{tabular}
    \label{tab:1.5b}
\end{table*}

\begin{table*}[!htbp]
    \centering
    \caption{Scaling results on the DeepScaleR-Preview-Dataset using the DeepSeek-R1-Distill-Qwen-7B backbone. }
    \renewcommand{\arraystretch}{1.1}
    \setlength{\tabcolsep}{5.5pt}
    \small 
    
    \begin{tabular}{@{}l c c c c c c c@{}}
        \toprule
        \textbf{} & AIME24 & AIME25 & MATH500 & AMC23 & Minerva & O-Bench & Avg. \\
        \midrule
        
        \multicolumn{8}{l}{\textit{\textbf{Models}}} \\ 
        \hspace{1em} DS-Distill-Qwen-7B\textsuperscript{*} & 0.402 & 0.292 & 0.873 & 0.688 & 0.355 & 0.471 & 0.513 \\

        \addlinespace[0.5ex] 
        \midrule
        
        \multicolumn{8}{l}{\textit{\textbf{Methods}}} \\
        \hspace{1em} GRPO\textsuperscript{*} & 0.498 & 0.394 & 0.916 & 0.807 & 0.381 & 0.555 & 0.592 \\
        \hspace{1em} DAPO\textsuperscript{*} & 0.454 & 0.335 & 0.907 & 0.799 & 0.388 & 0.535 & 0.570 \\
        \hspace{1em} TRPA\textsuperscript{*} & 0.510 & 0.367 & 0.898 & 0.779 & 0.379 & 0.534 & 0.578 \\
        \hspace{1em} GSPO & 0.525 & 0.396 & 0.926 & 0.852 & 0.402 & 0.588 & 0.615 \\
        \hspace{1em} REAL & \textbf{0.571} & \textbf{0.433} & \textbf{0.929} & \textbf{0.858} & \textbf{0.402} & \textbf{0.602} & \textbf{0.632} \\
        \bottomrule
    \end{tabular}
    \label{tab:7b}
\end{table*}

\subsection{Gradient analysis of REAL}

\textbf{Proposition 5.1.}
\label{Proposition 5.1}
(\emph{refer to Appendix~\ref{appendix:Gradient Analysis of REAL} for detailed proof}) 
For both positive and negative rollouts, REAL induces a \emph{\textbf{monotonic}} and \emph{\textbf{bounded}} gradient weighting that is upper-bounded by $\frac{1}{\tau}$. The influence of a rollout is smoothly modulated by its relative log-probability, yielding an implicit and adaptive form of gradient clipping. Specifically, for the rollout $o^k$, the gradient magnitude induced by REAL is given by:
\begin{equation}
\label{eq:real gradient}
|\mathcal{W}_{\text{REAL}}|
=
\begin{cases}
\dfrac{1}{\tau}\dfrac{1}{1 + C_+ e^{\bar{s}^k / \tau}}, & r = 1, \\[8pt]
\;\;\dfrac{1}{\tau}\dfrac{1}{1 + C_- e^{-\bar{s}^k / \tau}}, & r = 0,
\end{cases}
\end{equation}

where
\begin{equation} 
\begin{aligned} 
C_+ &= 1 + \sum_{i=1, i \neq k}^{|\mathcal{O}_+|} e^{-\bar{s}^i / \tau}  \\
C_- &= 1 + \sum_{j=1, j \neq k}^{|\mathcal{O}_-|} e^{\bar{s}^j / \tau}
\end{aligned} 
\end{equation}

$C_+\ge1$ and $C_-\ge1$ are context-dependent constants determined by other rollouts in the group.

As illustrated in Fig.~\ref{fig:intro-1}, Proposition~5.1 shows that REAL assigns a bounded gradient weight for all rollouts. For positive rollouts ($r=1$), the gradient magnitude monotonically decreases with increasing relative log-probability $\bar{s}^k$, preventing overly confident updates. For negative rollouts ($r=0$), rollouts with larger $\bar{s}^k$ are penalized more strongly. Crucially, in both cases the gradient magnitude is upper-bounded by $\frac{1}{\tau}$, ensuring that all updates remain within a controlled range.

\paragraph{Discussion.}
Taking advantage of the bounded gradient weighting induced by REAL, the framework naturally avoids overly large updates even without an explicit KL divergence regularization. In Sec.~\ref{sec:further analysis}, we empirically verify that REAL achieves comparable performance without any KL penalty, confirming that the adaptive gradient clipping mechanism is sufficient to ensure training stability.

\begin{table*}[!htbp]
    \centering
    \caption{Generalizability study on the DAPO-Math-17K dataset using the DeepSeek-R1-Distill-Qwen-1.5B model. }
    \renewcommand{\arraystretch}{1.1}
    \setlength{\tabcolsep}{5.5pt}
    \small
    
    \begin{tabular}{@{}l c c c c c c c@{}}
        \toprule
        \hspace{1em} \textbf{Methods} & AIME24 & AIME25 & MATH500 & AMC23 & Minerva & O-Bench & Avg. \\
        \midrule
        
        \hspace{1em} GRPO & 0.342 & 0.256 & 0.842 & 0.672 & 0.267 & 0.458 & 0.473 \\
        \hspace{1em} DAPO & 0.275 & 0.229 & 0.812 & 0.653 & 0.256 & 0.441 & 0.444 \\
        \hspace{1em} TRPA\textsuperscript{*} & 0.346 & 0.279 & 0.836 & 0.683 & 0.281 & 0.450 & 0.479 \\
        \hspace{1em} {GSPO}  & 0.365 & 0.290 & \textbf{0.867} & 0.736 & \textbf{0.313} & \textbf{0.499} & 0.511 \\
        \hspace{1em} {REAL}   & \textbf{0.402} & \textbf{0.306} & 0.866 & \textbf{0.759} & 0.312 & 0.495 & \textbf{0.523} \\

        \bottomrule
    \end{tabular}
    \label{tab:dapo17k}
\end{table*}

\section{Experiments}
\label{sec:exp}
\subsection{Setup}

\paragraph{Models and Baselines.}  We employ DeepSeek-R1-Distill-Qwen-1.5B and 7B as our base models. To evaluate the effectiveness of REAL, we compare it against several representative RLVR baselines: GRPO \cite{guo2025deepseek}, DAPO \cite{yu2025dapo}, TRPA \cite{su2025trust}, and GSPO \cite{zheng2025group}. Specifically, DAPO \cite{yu2025dapo} improves upon GRPO by introducing a clip higher strategy to mitigate entropy collapse and employing token-mean normalization to compute rewards over token-level averages. TRPA \cite{su2025trust} integrates rule-based optimization with preference-based optimization for reasoning tasks. GSPO \cite{zheng2025group} shifts the optimization to the sequence level for both importance sampling and advantage estimation to improve training stability.
\vspace{-1em}

\paragraph{Tasks and Datasets.} 
We evaluate REAL on a comprehensive suite of mathematical reasoning tasks. For training, we utilize the DeepScaleR-Preview-Dataset~\cite{deepscaler2025}, which comprises approximately 40.3k unique problem-answer pairs designed for reinforcement learning with verifiable rewards. Model performance is assessed across six diverse benchmarks: AIME 2024, AIME 2025, MATH 500~\cite{hendrycks2021measuring,lightman2023let}, AMC 2023, Minerva~\cite{lewkowycz2022solving}, and Olympiad Bench (O-Bench)\cite{he2024olympiadbench}. In line with standard evaluation protocols\cite{guo2025deepseek,deepscaler2025}, we report the $Pass@1$ metric, computed as the average accuracy over $k=16$ sampled responses per question.

\begin{figure*}[t]
    \centering
    \includegraphics[width=\textwidth]{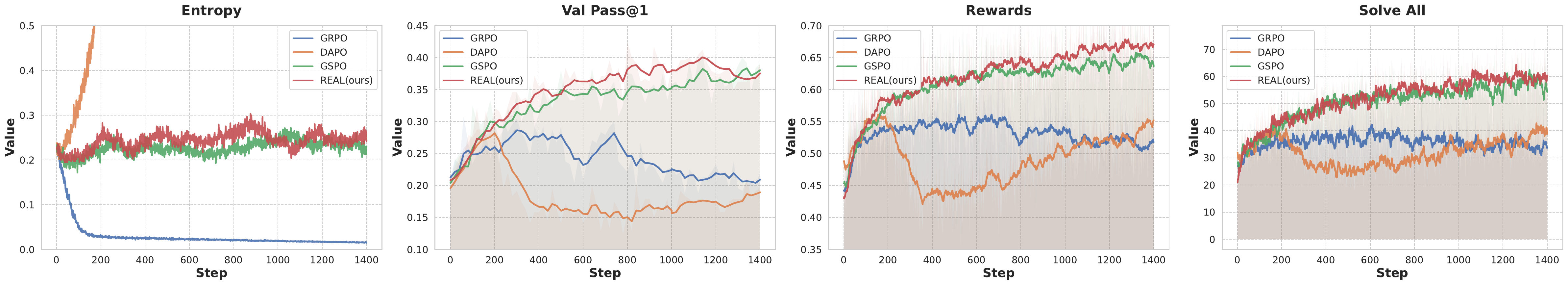} 
    \caption{\textbf{Training dynamics of the DeepSeek-R1-Distill-Qwen-1.5B model.} We compare \textbf{REAL} against three baselines—GRPO, DAPO, and GSPO—across four key metrics: Entropy, Validation Score (Pass@1 on AIME 2024, averaged over 8 samples),  Reward, and 100\% Solved Ratio.}
    \vspace{-1em}
    \label{fig:training_dynamics}
\end{figure*}

\paragraph{Implementation Details.} 
All models are optimized using AdamW with a weight decay of 0.01. We employ a global batch size of 128 (mini-batch size 32), with $G=8$ rollouts sampled per prompt. The sampling temperature is 0.6 for both training and inference, and the maximum context length is set to 8,192 tokens. For the 1.5B model, we use a learning rate of $2 \times 10^{-6}$ and train for 1,400 steps; for the 7B model, the learning rate is $1 \times 10^{-6}$ over 1,000 steps.

Method-specific hyperparameters are as follows: for GRPO, we set the KL penalty coefficient to $\beta=0.001$. For DAPO, we adopt the asymmetric clipping thresholds $\varepsilon_{\text{low}}=0.2$ and $\varepsilon_{\text{high}}=0.28$. For GSPO, the sequence-level clipping thresholds are set to $\varepsilon_{\text{low}}=3 \times 10^{-4}$ and $\varepsilon_{\text{high}}=4 \times 10^{-4}$. Regarding TRPA, we adopt the results reported in \citet{li2025disco}. For our proposed REAL, the temperature parameter is fixed at $\tau=0.5$. Notably, to isolate the impact of the core objectives, REAL, DAPO and GSPO baselines are trained without an explicit KL penalty term.
\subsection{Main Results}

\paragraph{Performance Analysis.}
REAL exhibits significant performance gains across all six mathematical reasoning benchmarks. As reported in Tables \ref{tab:1.5b} and \ref{tab:7b}, REAL consistently outperforms all baseline methods across different model scales. For the 1.5B model, REAL achieves a substantial lead over GRPO and DAPO, improving the average Pass@1 by 9.5 and 6.7 points, respectively. Even when compared to GSPO, a strongly competitive baseline, REAL maintains a steady advantage, particularly on challenging benchmarks like AIME 2024 and AIME 2025. Scaling to the 7B model, REAL achieves an average Pass@1 of 63.2\%, surpassing GSPO by 1.7 points. These improvements can be directly attributed to REAL’s ability to rectify gradient misassignment and gradient domination, ensuring that learning signals are more effectively allocated.

To further verify the robustness of our framework, we evaluated REAL on the DAPO-Math-17K dataset. As shown in Table \ref{tab:dapo17k}, REAL continues to surpass other methods, maintaining its performance edge despite the change in data distribution. This consistency confirms REAL's potential as a more principled and generalizable approach to RLVR.

\paragraph{Training Stability.}
Fig.~\ref{fig:training_dynamics} shows the training dynamics of the 1.5B model across entropy, validation Pass@1 on AIME 2024, and rewards. GRPO suffers from entropy collapse, while DAPO exhibits entropy explosion, both leading to poor validation performance. In contrast, REAL maintains stable entropy throughout the 1,400 steps. This stable entropy profile suggests a well-controlled exploration–exploitation trade-off and translates to consistent improvements: REAL achieves steady growth in both training rewards and validation Pass@1 scores, ultimately outperforming all baselines. These results demonstrate that REAL provides more reliable and effective optimization compared to standard GRPO-style methods.

\begin{figure}[t]
\centering
\includegraphics[width=\columnwidth]{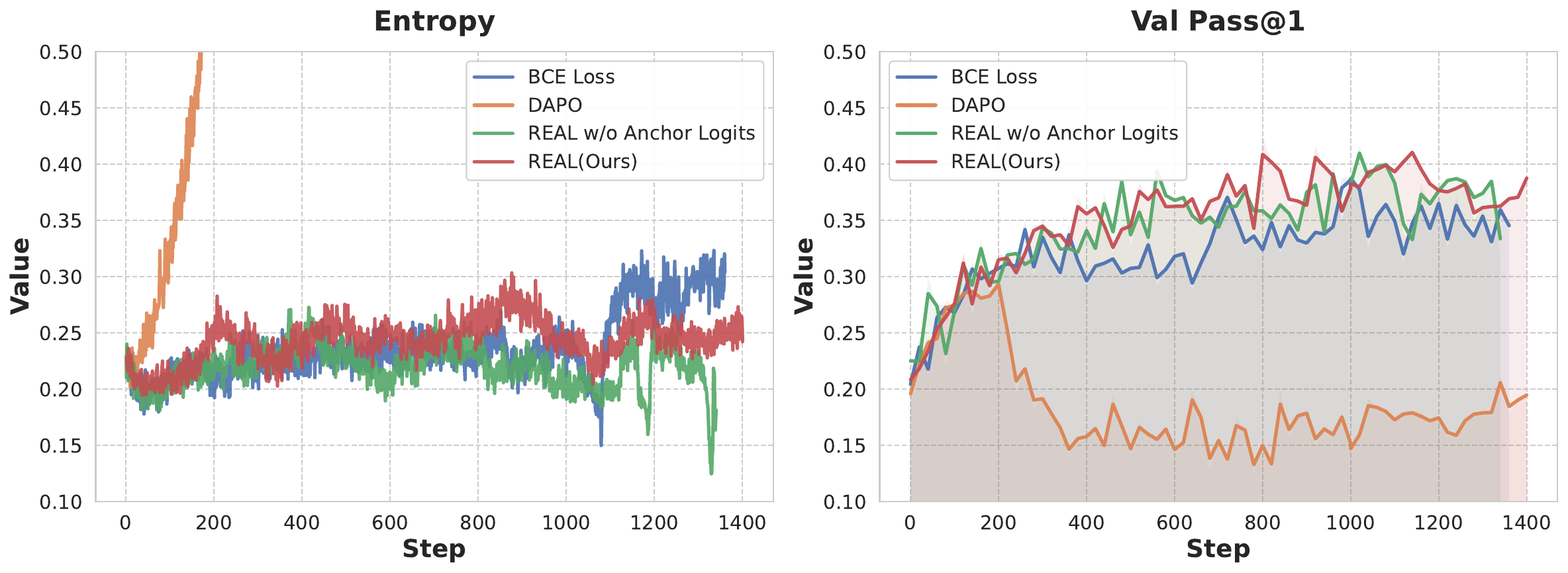}
\caption{\textbf{Further analysis on BCE loss and Anchor Logits.} Training dynamics and validation Pass@1 on AIME 2024 of different methods: DAPO, REAL w/ BCE Loss, REAL w/o Anchor Logits, and \textbf{REAL (Ours)}.}
\label{fig:score_ablation}
\end{figure}

\subsection{Further Analysis}
\label{sec:further analysis}
\paragraph{Comparison with BCE Classification Loss.}

We investigate whether REAL can be extended to other classification objectives by replacing the softmax-based loss with Binary Cross-Entropy (BCE). For this variant, we use the sequence-level average log-ratio $\bar{s}$ as defined in Eq.~\ref{eq:relative log-prob}. In RLVR, the verifiable reward $r \in \{0, 1\}$ naturally serves as the binary classification label. The BCE objective is formulated as:
\begin{align}
\label{eq:BCE_Unified}
& \mathcal{L}_{\mathrm{BCE}} = \mathcal{L}_+ + \mathcal{L}_- \notag \\
& = - \left(
    \sum_{\mathcal{O}_+} \log \sigma(\bar{s}_i/\tau) 
    + 
    \sum_{\mathcal{O}_-} \log (1 - \sigma(\bar{s}_j/\tau)) 
\right)
\end{align}

As shown in Table~\ref{tab:ablation}, the BCE variant achieves 50.4\% Pass@1, which is lower than REAL's 52.6\%. We attribute this to two factors: 1) the softmax loss naturally encourages group-wise competition among rollouts, aligning gradient allocation with relative rollout quality, and 2) our Anchor Logits mechanism stabilizes training by providing a consistent reference for the relative log-probabilities, which is not present in the BCE formulation.

\paragraph{Necessity of KL Divergence.}  
Previous works employ a KL Divergence loss to stabilize training \cite{guo2025deepseek}. We investigate its necessity in REAL under three settings: 1) REAL without KL Divergence, 2) REAL with KL Divergence computed between the current policy and a reference model, and 3) REAL with KL Divergence computed between the current policy and the old policy.  
As shown in Tab.~\ref{tab:ablation}, there is no significant performance difference among these settings, suggesting that REAL inherently provides an implicit and adaptive form of gradient clipping, making explicit KL loss largely unnecessary.

\paragraph{Effect of Anchor Logits.}  
We also evaluate REAL without Anchor Logits, as defined in Eq. \ref{eq:vanilla loss}. As shown in Fig. \ref{fig:score_ablation}, while REAL w/o Anchor Logits remains stable and achieves competitive performance, it is slightly outperformed by the one with Anchor Logits. This indicates Anchor Logits provide an additional benefit by further stabilizing training and providing a clearer optimization direction.

\paragraph{Impact of Hyperparameter $\tau$.} We analyze how the temperature  $\tau$ affects training stability and performance. As shown in Tab.~\ref{tab:ablation}, we evaluate $\tau \in \{0.1, 0.5, 1.0\}$. We find that $\tau=0.1$ makes the training unstable and the model collapse early. In contrast, $\tau=0.5$ and $\tau=1.0$ remain stable throughout training.
These observations are consistent with the gradient analysis of REAL in Proposition 5.1. Specifically, REAL induces a monotonic and bounded gradient weighting, upper-bounded by $1/\tau$. A lower temperature increases this upper bound, which can lead to training instability. In our experiments, we use $\tau=0.5$ for all datasets.

\section{Related Works}
\subsection{Large Reasoning Models and RLVR}

Recent advances in LLMs, such as OpenAI's o-series~\cite{openo1}, DeepSeek-R1~\cite{guo2025deepseek}, Kimi K2~\cite{team2025kimi}, and the Gemini series~\cite{comanici2025gemini}, have demonstrated remarkable reasoning capabilities on complex tasks. These successes highlight the potential of reinforcement learning as a post-training strategy to further enhance reasoning performance.
Reinforcement Learning with Verifiable Rewards has emerged as the dominant paradigm for post-training optimization of reasoning models. RLVR simplifies reward design by leveraging rule-based verification functions (e.g., exact answer matching) to generate reward signals, eliminating the need for separate reward models and thereby substantially reducing memory and computational overhead during RL training.

Building upon DeepSeek-R1's core algorithm, GRPO~\cite{shao2024deepseekmath}, a number of RLVR variants have been proposed, focusing primarily on algorithmic improvements~\cite{yu2025dapo, liu2025understanding, chu2025gpg, lin2025cppo, team2025kimi, su2025trust, cui2025entropy, zhu2025flowrl}, reward design~\cite{cheng2025reasoning, wu2025quantile}, and sampling strategies~\cite{yu2025dapo, skywork-or1-2025, zhang2025srpo, hu2025open}.  

Despite these empirical successes, most variants are still built upon GRPO, which suffers from inherent gradient issues. In this work, we identify two fundamental gradient mismatches, Gradient Misassignment in Positives and Gradient Domination in Negatives, that lead to inefficient credit assignment and suboptimal policy updates, motivating the development of our proposed approach.

\begin{table}[t]
\centering
\caption{Further Analysis results for BCE loss, Anchor Logits, temperature $\tau$ and KL Divergence settings. Averaged Pass@1 (\%) across six mathematical reasoning benchmarks is reported.}
\label{tab:ablation}
\resizebox{0.27\textwidth}{!}{
\begin{tabular}{lcc}
\toprule
 \textbf{Settings} & \textbf{Pass@1 (\%)} \\
 \midrule
w/ BCE loss & 50.4 \\
w/o Anchor Logits & 51.4 \\
 \midrule
$\tau = 0.1$ & 52.3 \\
$\tau = 1.0$ & 51.5 \\
 \midrule
KL w/ reference model & 52.1 \\
KL w/ old policy & 51.9 \\
\midrule
REAL & 52.6 \\
\bottomrule
\end{tabular}
}
\vspace{-1em}
\end{table}

\subsection{Classification Objectives}
Classification is a long-standing and fundamental problem in machine learning. Numerous approaches have been proposed to achieve this objective. Among them, softmax cross-entropy loss remains the most fundamental and widely used due to its simplicity and effectiveness. Building on softmax cross-entropy, a variety of variants, such as NormFace \cite{wang2017normface}, CosFace \cite{wang2018cosface}, ArcFace \cite{deng2019arcface}, and Circle Loss\cite{sun2020circle}, have been developed to further improve intra-class compactness and inter-class separability. These methods are particularly effective in challenging scenarios, including face recognition and fine-grained image classification.

Our REAL framework relies solely on the softmax cross-entropy loss, leveraging its simplicity and effectiveness without any extra enhancements.

\section{Conclusions}
We identify two gradient mismatches in the GRPO-style RLVR methods, Gradient Misassignment in Positives and Gradient Domination in Negatives, which lead to inefficient and suboptimal policy updates. To address these issues, we propose REAL, a novel framework that reframes verifiable rewards as categorical labels and formulates policy optimization as a classification problem. This design yields balanced and bounded gradient allocation across rollouts, improving training stability and policy performance. This paper provides new insights into RLVR optimization and establishes classification reformulation as a principled path toward stable and effective policy learning.

\section*{Impact Statement}
Reliable and stable optimization is essential for the responsible development of large language models. This work identifies two fundamental gradient mismatches in existing reinforcement learning with verifiable rewards (RLVR) methods, which can lead to inefficient training dynamics and unpredictable model behavior. By reformulating verifiable rewards as categorical labels, our proposed REAL framework provides a more principled and stable optimization paradigm, enabling the development of more transparent, robust, and trustworthy AI systems.

\bibliography{example_paper}
\bibliographystyle{icml2026}

\newpage
\appendix
\onecolumn

\section{Proof of Propositions}
\subsection{Gradient Analysis of GRPO}
\label{appendix:Gradient Analysis of GRPO}

Formally, the gradient of GRPO with respect to its parameter $\theta$ is given as follows:

\begin{equation}
\begin{aligned}
\nabla_\theta \mathcal{J}_{\text{GRPO}} &= \mathbb{E}_{} \Bigg[ \nabla_\theta \Big( \frac{1}{|o|} \sum_{t=1}^{|o|} \min(\rho_t^{\theta} A_t,  \text{clip}(\rho_t^\theta, 1-\varepsilon, 1+\varepsilon)A_t) \Big) \Bigg] \\
&= \mathbb{E}_{} \left[ \frac{1}{|o|} \sum_{t=1}^{|o|} \mathbb{I}_{\text{clip}} \cdot A_t \nabla_\theta \frac{\pi_\theta(o_t|q)}{\pi_{\text{old}}(o_t|q)} \right] \\
&= \mathbb{E}_{} \left[ \frac{1}{|o|} \sum_{t=1}^{|o|} \mathbb{I}_{\text{clip}} \cdot A_t  \rho_t^\theta \nabla_\theta \log \pi_\theta(o_t|q) \right] \\
&= \mathbb{E}_{} \left[ \frac{1}{|o|} \sum_{t=1}^{|o|} \mathbb{I}_{\text{clip}} \cdot A_t  e^{s_t} \nabla_\theta \log \pi_\theta(o_t|q) \right],
\end{aligned}
\end{equation}

where $\mathbb{I}_{clip}$ is the indicator function for the clipping constraint. $s_t$ is defined as the relative logarithmic probability between $\pi_\theta(o_t|q)$ and $\pi_{\text{old}}(o_t|q)$:
\begin{equation}
\label{eq:s_definition_app}
\begin{aligned}
s_t = \log \frac{\pi_\theta(o_t|q)}{\pi_{\text{old}}(o_t|q)}
\end{aligned}
\end{equation}

Then, for the $t$-th token, the absolute gradient magnitude could be rewritten as:

\begin{equation}
|\mathcal{W}_\text{GRPO}| = 
\begin{cases} 
|A \cdot e^{s_t}|, & \text{if } \mathbb{I}_{\text{clip}}=1  \\
0, & \text{otherwise}.
\end{cases}
\end{equation}

\subsection{Gradient Analysis of REAL}
\label{appendix:Gradient Analysis of REAL}

We analyze the gradient magnitude of REAL with respect to the policy parameters~$\theta$.
Formally, the gradient of the REAL objective is given by
\begin{equation}
\begin{aligned}
\nabla_\theta \mathcal{L}_{\text{REAL}}
=
\mathbb{E}\Bigg[
\nabla_\theta \log \Bigl( 1 + \sum_{\mathcal{O}_+} e^{-\bar{s}^i / \tau} \Bigr)
+
\nabla_\theta \log \Bigl( 1 + \sum_{\mathcal{O}_-} e^{\bar{s}^j / \tau} \Bigr)
\Bigg].
\end{aligned}
\end{equation}

The normalized relative log-probability score for rollout $o^k$ is defined as
\begin{align}
\bar{s}^k
&=
\frac{1}{|o^k|}
\sum_{t=1}^{|o^k|}
\log
\frac{\pi_\theta(o_t^k \mid q)}
{\pi_{{\text{old}}}(o_t^k \mid q)}
=
\frac{1}{|o^k|}
\sum_{t=1}^{|o^k|}
s_t^k .
\end{align}

\paragraph{Positive samples ($k \in \mathcal{O}_+$).}
We first consider the contribution of a positive rollout $k \in \mathcal{O}_+$:
\begin{equation}
\begin{aligned}
\nabla_\theta \mathcal{L}_{\text{REAL}}
&=
\mathbb{E}
\Bigg[
\nabla_\theta
\log
\Bigl( 1 + \sum_{\mathcal{O}_+} e^{-\bar{s}^i / \tau} \Bigr)
\Bigg] \\
&=
\mathbb{E}
\Bigg[
\frac{1}{1 + \sum_{\mathcal{O}_+} e^{-\bar{s}^i / \tau}}
\cdot
\left(
-\frac{1}{\tau} e^{-\bar{s}^k / \tau}
\right)
\nabla_\theta \bar{s}^k
\Bigg].
\end{aligned}
\end{equation}

Since
\begin{equation}
\nabla_\theta \bar{s}^k
=
\frac{1}{|o^k|}
\sum_{t=1}^{|o^k|}
\nabla_\theta \log \pi_\theta(o_t^k \mid q),
\end{equation}
we obtain
\begin{equation}
\begin{aligned}
\nabla_\theta \mathcal{L}_{\text{REAL}}
=
\mathbb{E}
\Bigg[
\frac{1}{|o^k|}
\sum_{t=1}^{|o^k|}
\left(
-\frac{1}{\tau}
\frac{e^{-\bar{s}^k / \tau}}
{1 + \sum_{i \in \mathcal{O}_+} e^{-\bar{s}^i / \tau}}
\right)
\nabla_\theta \log \pi_\theta(o_t^k \mid q)
\Bigg].
\end{aligned}
\end{equation}

Hence, the absolute magnitude  induced by REAL is
\begin{equation}
\bigl|\mathcal{W}_{\text{REAL}}\bigr|
=
\frac{1}{\tau}
\frac{e^{-\bar{s}^k / \tau}}
{1 + \sum_{\mathcal{O}_+} e^{-\bar{s}^i / \tau}}.
\end{equation}

Separating rollout $k$ from the remaining positive samples yields
\begin{equation}
1 + \sum_{i=1}^{|\mathcal{O}_+|} e^{-\bar{s}^i / \tau}
=
\underbrace{
1 + \sum_{i=1, i \neq k}^{|\mathcal{O}_+|} e^{-\bar{s}^i / \tau}
}_{C}
+
 e^{-\bar{s}^k / \tau}.
\end{equation}

Substituting these expressions gives
\begin{equation}
\begin{aligned}
\bigl|\mathcal{W}_{\text{REAL}}\bigr|
&=
\frac{1}{\tau}
\frac{1}{1 + C e^{\bar{s}^k / \tau}}.
\end{aligned}
\end{equation}

The derivation for negative samples follows analogously.

Combining positive and negative samples, the absolute gradient magnitude of REAL can be summarized as
\begin{equation}
|\mathcal{W}_{\text{REAL}}|
=
\begin{cases}
\dfrac{1}{\tau}\dfrac{1}{1 + C_+ e^{\bar{s}^k / \tau}}, & r = 1, \\[8pt]
\;\;\dfrac{1}{\tau}\dfrac{1}{1 + C_- e^{-\bar{s}^k / \tau}}, & r = 0,
\end{cases}
\end{equation}
where
\begin{equation} 
\begin{aligned} 
C_+ &= 1 + \sum_{i=1, i \neq k}^{|\mathcal{O}_+|} e^{-\bar{s}^i / \tau}  \\
C_- &= 1 + \sum_{j=1, j \neq k}^{|\mathcal{O}_-|} e^{\bar{s}^j / \tau}
\end{aligned} 
\end{equation}

$C_+\ge1$ and $C_-\ge1$ are context-dependent constants determined by other rollouts in the group.

\end{document}